\documentclass[acmtog]{acmart}

\usepackage{booktabs} 

\usepackage{bm}
\usepackage{amsmath}
\usepackage{amsfonts}
\usepackage{makecell}
\usepackage{CJKutf8}
\usepackage[utf8]{inputenc}

\usepackage{xspace}
\usepackage{xcolor}

\newcommand{\eg}{e.g.\xspace}

\usepackage{amsmath,amsfonts}

\usepackage{bbding}
\usepackage{multirow}

\usepackage{algorithmic}
\usepackage{algorithm}

\acmJournal{TOG}

\DeclareUnicodeCharacter{2194}{\ensuremath{\leftrightarrow}}

\copyrightyear{2025}
\acmYear{2025}
\setcopyright{acmlicensed}\acmConference[SIGGRAPH Conference Papers '25]{Special
Interest Group on Computer Graphics and Interactive Techniques Conference
Conference Papers }{August 10--14, 2025}{Vancouver, BC, Canada}
\acmBooktitle{Special Interest Group on Computer Graphics and Interactive
Techniques Conference Conference Papers (SIGGRAPH Conference Papers '25), August
10--14, 2025, Vancouver, BC, Canada}
\acmDOI{10.1145/3721238.3730731}
\acmISBN{979-8-4007-1540-2/2025/08}

\begin{document}

\title{AssetDropper: Asset Extraction via Diffusion Models with Reward-Driven Optimization}

\author{Lanjiong Li}
\email{lli746@connect.hkust-gz.edu.cn}
\authornote{Equal contribution}
\orcid{0009-0008-6738-7066}
\affiliation{%
  \institution{The Hong Kong University of Science and Technology (Guangzhou)}
  \city{Guangzhou}
  \country{China}
}

\author{Guanhua Zhao}
\email{2301212814@stu.pku.edu.cn}
\authornotemark[1]
\orcid{0009-0003-2620-9937}
\affiliation{%
  \institution{School of Electronic and Computer Engineering, Peking University}
  \city{Shenzhen}
  \country{China}
}

\author{Lingting Zhu}
\email{ltzhu99@connect.hku.hk}
\authornotemark[1]
\authornote{Project lead}
\orcid{0000-0002-1478-3232}
\affiliation{%
  \institution{The University of Hong Kong}
  \city{Hong Kong}
  \country{China}
}

\author{Zeyu Cai}
\email{zcai701@connect.hkust-gz.edu.cn}
\orcid{0009-0006-5422-4044}
\affiliation{%
  \institution{The Hong Kong University of Science and Technology (Guangzhou)}
  \city{Guangzhou}
  \country{China}
}

\author{Lequan Yu}
\email{lqyu@hku.hk}
\orcid{0000-0002-9315-6527}
\affiliation{%
  \institution{The University of Hong Kong}
  \city{Hong Kong}
  \country{China}
}

\author{Jian Zhang}
\email{zhangjian.sz@pku.edu.cn}
\orcid{0000-0001-5486-3125}
\affiliation{%
  \institution{School of Electronic and Computer Engineering, Peking University}
  \city{Shenzhen}
  \country{China}
}

\author{Zeyu Wang}
\email{zeyuwang@ust.hk}
\authornote{Corresponding author} 
\orcid{0000-0001-5374-6330}
\affiliation{%
  \institution{The Hong Kong University of Science and Technology (Guangzhou)}
  \city{Guangzhou}
  \country{China}
}
\affiliation{%
  \institution{The Hong Kong University of Science and Technology}
  \city{Hong Kong}
  \country{China}
}

\begin{abstract}
Recent research on generative models has primarily focused on creating product-ready visual outputs; however, designers often favor access to standardized asset libraries, a domain that has yet to be significantly enhanced by generative capabilities.
Although open-world scenes provide ample raw materials for designers, efficiently extracting high-quality, standardized assets remains a challenge.
To address this, we introduce AssetDropper, the first generative framework designed to extract any asset from reference images, providing artists with an open-world asset palette. Our model adeptly extracts a front view of selected subjects from input images, effectively handling complex scenarios such as perspective distortion and subject occlusion.
We establish a synthetic dataset of more than 200,000 image-subject pairs and a real-world benchmark with thousands more for evaluation, facilitating the exploration of future research in downstream tasks.
Furthermore, to ensure precise asset extraction that aligns well with the image prompts, we employ a pre-trained reward model to achieve a closed loop with feedback. We design the reward model to perform an inverse task that pastes the extracted assets back into the reference sources, which assists training with additional consistency and mitigates hallucination.
Extensive experiments show that, with the aid of reward-driven optimization, AssetDropper achieves the state-of-the-art results in asset extraction.
Our code and dataset are available at AssetDropper.github.io.

\end{abstract}

%
%


%
%


\begin{teaserfigure}
  \includegraphics[width=\textwidth]{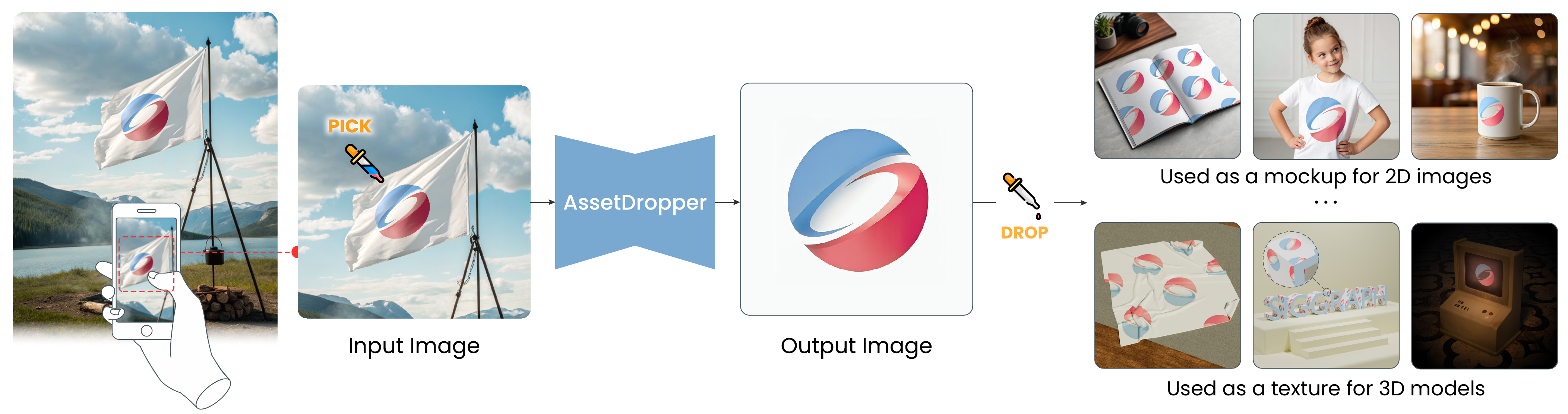}
  \caption{The application scenarios of AssetDropper, a novel model designed to extract assets from user-specified image regions. The extracted assets can seamlessly be applied to various downstream tasks in design and visualization workflows, e.g., 2D mockup creation and 3D model texturing.}
  \label{fig:teaser}
\end{teaserfigure}

\maketitle

\section{Introduction}
\begin{figure}[htbp]
    \centering
    \includegraphics[width=0.45\textwidth]{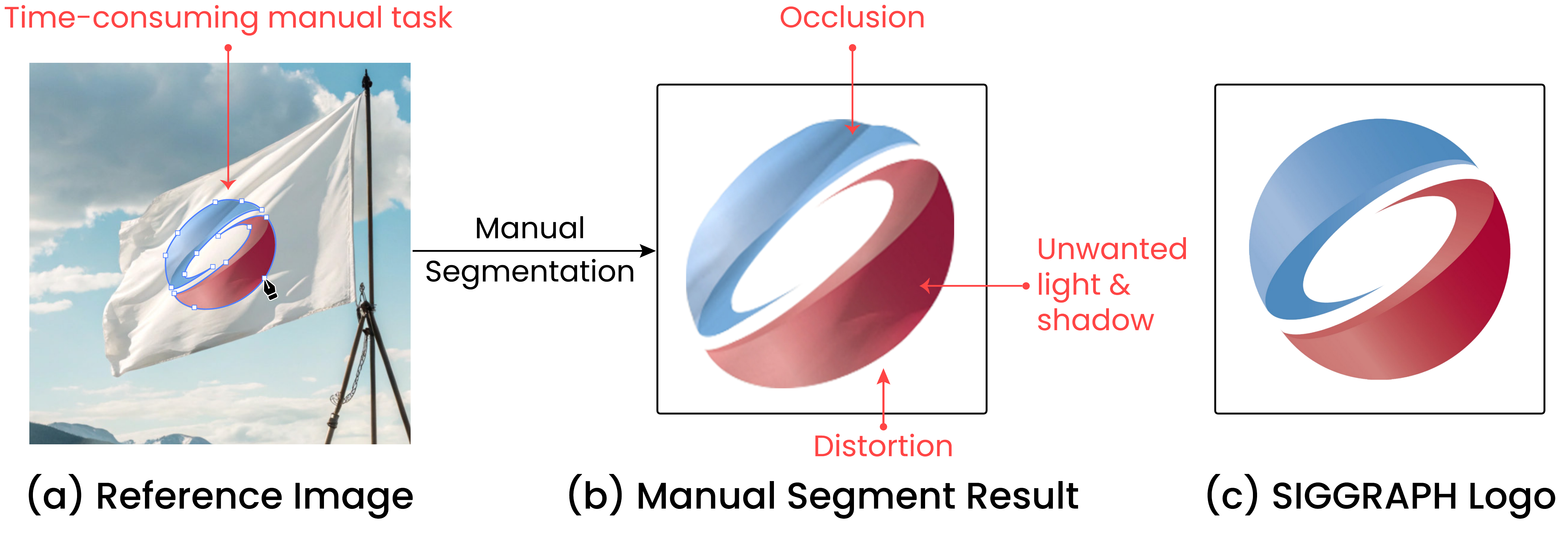}
    \caption{Illustration of the manual segmentation workflow. Users manually draw Bezier curves using the pen tool to segment the asset.}
    \label{fig:photoshop}
\end{figure}

Generative methods, particularly those based on diffusion models~\cite{ho2020denoising,song2020denoising,rombach2022high,tewel2024trainingfreeconsistenttexttoimagegeneration,hertz2024stylealignedimagegeneration,liu2024intelligentgrimmopenended}, have achieved remarkable success in translating the vivid imaginations of users into tangible visual outputs. However, much existing research has concentrated on producing ready-to-use visual outputs, assisted with controllable branches that handle image prompts~\cite{zhang2023adding, mou2024t2i, ye2023ip}. Despite these advances, these methods fall short in meeting the practical needs of user-driven creation within real-world workflows. 
Standardized asset libraries support designers by providing versatile resources applicable across various fields, including advertising, game development, and educational content creation. However, integrating generative models with standardized asset libraries remains an area under active exploration. Currently, if users wish to extract assets from real-world images for their subsequent design needs, they rely on traditional software tools like Photoshop for tasks such as image editing and segmentation. This process is not only labor-intensive but also heavily reliant on the skills of the artist, often resulting in content of lower quality characterized by distortions and occlusions, as illustrated in Fig.~\ref{fig:photoshop}. These issues necessitate time-consuming post-processing, underscoring the need for more efficient and user-friendly asset extraction methods. 

Existing generative AI techniques fail to achieve effective asset extraction, primarily due to their limited control capabilities. While diffusion models enhanced with ControlNet~\cite{zhang2023adding} or IP-Adapter~\cite{ye2023ip} can generate images aligned with spatial, style, or subject controls, such guidance remains insufficient for practical asset extraction, particularly in scenarios involving distortions and occlusions commonly encountered in asset extraction tasks. An alternative approach is to frame asset extraction as an image editing problem, where the source image is modified to produce the desired content. Recent advances in universal generation methods, such as InstructPix2Pix~\cite{brooks2023instructpix2pix} and OmniGen~\cite{xiao2024omnigen}, shed light on any-form image creation based on human instructions. However, their performance remains unsatisfactory for our specific tasks, underscoring the need for a tailored solution.

We approach this task by developing a task-specific diffusion model tailored for asset extraction, which takes the source image and its corresponding mask as input. This is a relatively feasible and somewhat mature solution, as it builds upon a growing body of research in image customization~\cite{chen2024anydoorzeroshotobjectlevelimage} and image-based virtual try-on (VTON)~\cite{choi2024improvingdiffusionmodelsauthentic,choi2021viton, ge2021parser, issenhuth2020not, morelli2023ladi}, which can be viewed as the inverse of our target task. While extensive engineering efforts can lead to promising results, the performance remains far from perfect, primarily due to two key challenges: 1) the inherent difficulty of the task, as the assets of interest may appear in arbitrary locations with varying distortions within the source image, and 2) the extracted asset often fails to align with human intentions, partly due to the high randomness of the model and the lack of consistency regularization.

Accordingly, we introduce several designs specifically tailored to address these two challenges. 1) The dataset issue is critical, as it significantly impacts the effectiveness and generalizability of the model. However, obtaining paired data of source images and their corresponding standardized assets is challenging and expensive. While we can repurpose VTON datasets by treating them as inverse tasks, they are limited to human clothing and fail to cover the diversity of standardized assets encountered in real-world scenarios. To address this, we developed a large-scale synthetic dataset including 200k pairs of images and standardized assets. Specifically, we collected over 10k standardized assets and simulated a realistic lighting environment in Blender. To emulate various extraction scenarios, we projected these assets onto commonly used meshes from daily life and rendered images from multiple angles, thereby constructing a comprehensive paired dataset, assisted with VTON datasets for enhancing generalizability. 2) Aligning generative models with human intentions is a key focus. Recent studies~\cite{liang2024richhumanfeedbacktexttoimage,li2024controlnetimprovingconditionalcontrols} propose post-training approaches to fine-tune models to better align with human preferences. A major challenge in asset extraction is the lack of consistency, which can result in outputs that deviate significantly from expectations. To address this, we adopt a reward-driven optimization method to enhance consistency during model training. We hypothesize that an extracted asset is ``good'' if it can be seamlessly reattached to the source image using a mask, facilitated by another generative model. Our design is inspired by ControlNet++~\cite{li2024controlnetimprovingconditionalcontrols}, which employs a discriminative model to improve condition consistency. However, a key distinction in our approach is the absence of a discriminative model. Instead, we first propose to use a generative model to provide feedback supervision, enabling more effective alignment with human intentions.

\begin{figure*}[t]
    \centering
    \includegraphics[width=0.9\textwidth]{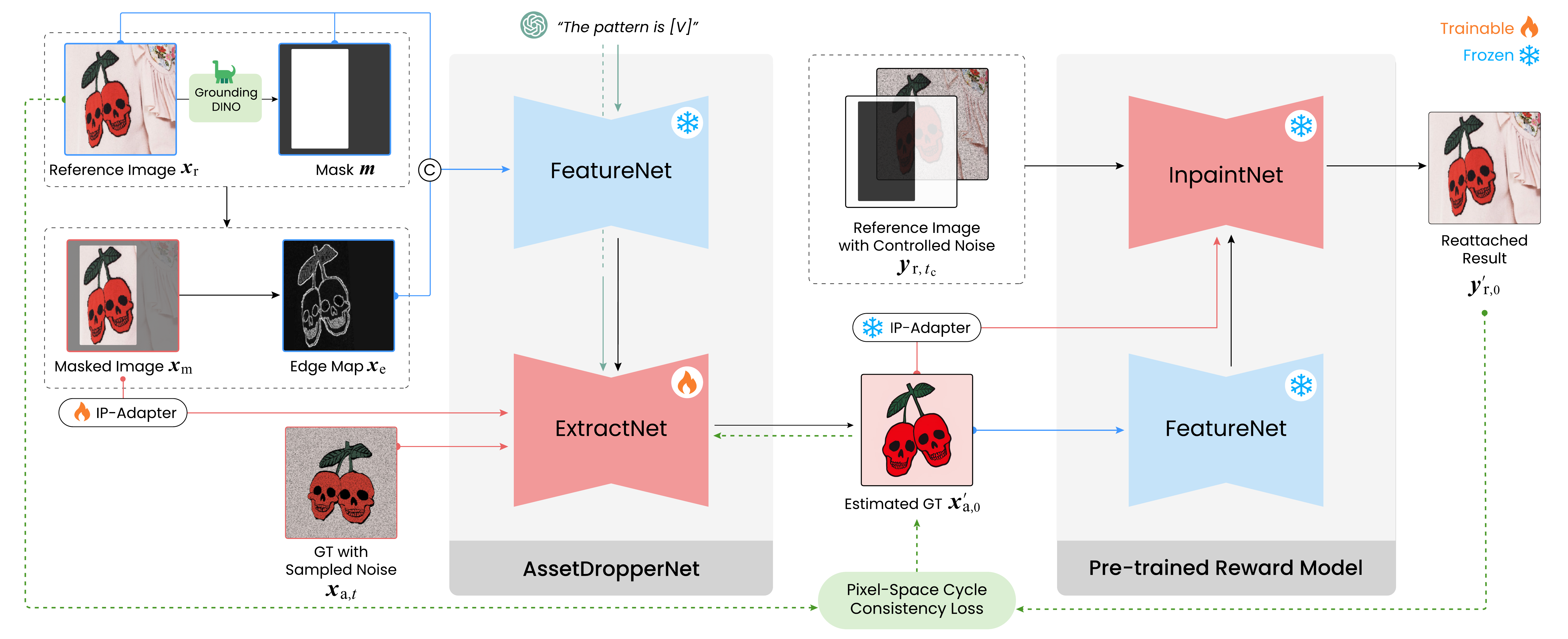}
    \caption{Overview of the AssetDropper framework. For the input reference image $\bm x_\text{r}$, we obtain the asset's mask $m$ through Grounding DINO, and further obtain the masked image $\bm x_\text{m}$ and the edge map $\bm x_\text{e}$ of the asset through the mask. The corresponding standardized asset is denoted by $\bm x_\text{a}$. Networks with the same color represent the same model (blue for SDXL UNet, red for SDXL-Inpainting UNet). We first use FeatureNet, a UNet that encodes the low-level features of the input $\bm x_\text{r}$, $\bm m$, and $\bm x_\text{e}$.
    We then use IP-Adapter to encode high-level semantics of the masked image $\bm x_\text{m}$. ExtractNet is the main UNet that serves as our generator, processing noisy data $\bm x_{\text{a},t}$. A detailed text prompt of the asset is provided by GPT-4o~\cite{gpt4o} for both FeatureNet and ExtractNet, where [V] denotes the subject, e.g., ``a surreal fusion of cherries and skulls, blending natural and macabre elements.'' We train a model with the same architecture as AssetDropperNet to perform the inverse task, reattaching the extracted asset back to the reference image while masking the inpainted area. This model is then used as our pre-trained reward model.}
    \label{fig:pipeline}
\end{figure*}

This paper explores high-quality and high-fidelity asset extraction from reference images, as illustrated in Fig.~\ref{fig:teaser}. Given a photo captured via a mobile device, users can instantly obtain a ready-to-use asset for their creative projects by simply selecting a region of interest. This approach offers a more efficient solution that seamlessly integrates into real-world creation workflows. Drawing an analogy to the dropper tool in Photoshop, we name our method AssetDropper, reflecting its capability to extract desired assets directly from in-the-wild images.

This paper makes the following contributions:
\begin{itemize}
    \item We propose the first generative framework, AssetDropper, to extract any asset with high quality and fidelity from reference images. Experimental results demonstrate that our method achieves state-of-the-art performance.
    \item We construct a Standardized Asset Palette Dataset and a benchmark for the asset extraction task, exploring a new AI-assisted workflow for user-driven creation.
    \item We introduce a reward-driven optimization method for AssetDropper to enhance consistency and align the generated assets with human preferences. To the best of our knowledge, this is the first work to employ a generative model as the reward model for image-based diffusion models.

\end{itemize}
\section{Related Work}
\subsection{Diffusion-Based Generative Models}
Diffusion models~\cite{sohl2015deep, song2019generative, ho2020denoising} have achieved great success in visual content generation.
Although text-to-image (T2I) diffusion models possess powerful capabilities to generate high-quality images from text prompts, natural language inherently lacks the fine-grained control needed for detailed image synthesis. To address this limitation, various methods have been proposed to integrate conditional control into T2I diffusion models~\cite{ho2022classifier, zhang2023adding, gal2022image, kumari2023multi, zhang2023controllable, chen2024training, qin2023unicontrolunifieddiffusionmodel, wang2024instancediffusion}. 
In this paper, we customize an image-conditioned diffusion model for extracting assets from reference images and demonstrate that it significantly enhances adaptability to real-world scenarios. This approach allows for more effective asset extraction that meets the specific requirements and variations encountered in practical applications.

\subsection{Image-Based Virtual Try-On}
Virtual try-on technology focuses on generating realistic images of a target individual wearing a specific garment through image-based techniques. Early approaches like GAN-based virtual try-on~\cite{choi2021viton, ge2021parser, lee2022high, xie2023gp} utilize a two-stage strategy: first deforming the garment to conform to the target person's body and then fusing the warped garment with the individual's image using a try-on generator. 
These approaches, which often leverage dense flow maps~\cite{zhou2016view} for precise deformation and employ techniques such as normalization or distillation to address misalignment~\cite{choi2021viton, ge2021parser, issenhuth2020not}, still face limitations when handling arbitrary human images with complex backgrounds or intricate poses. With the success of diffusion models in generative tasks~\cite{ho2020denoising, song2020score}, recent research has increasingly resorted to these models for virtual try-on applications. For instance, TryOnDiffusion~\cite{zhu2023tryondiffusion} utilizes two parallel UNet networks for garment transformation, demonstrating the potential of diffusion models in this domain. 
Subsequent research has approached virtual try-on as an image inpainting problem based on examples~\cite{yang2023paint}. These studies fine-tune inpainting diffusion models~\cite{kim2024stableviton, morelli2023ladi, choi2024improvingdiffusionmodelsauthentic} on virtual try-on datasets to generate high-quality images for virtual try-on applications. However, these methods still face challenges such as significant loss of spatial information and inaccuracies in garment warping. These issues are consistent with the challenges faced in our asset extraction task, which requires the effective and precise handling of the spatial and geometric information of the objects from which the assets are extracted.

\subsection{Visual Reward Models}
Diffusion-based models are generally trained on extensive datasets, which endow them with robust generalization capabilities. However, to tailor these models to specific tasks or to meet precise human-centric requirements, post-training adjustments are often necessary. This process enables them to respond accurately to human-level instructions for versatile content generation~\cite{brooks2023instructpix2pix,lin2024pixwizard}.
Beyond simple fine-tuning of existing and additional branches, methods that incorporate reward-based objectives are garnering significant attention~\cite{prabhudesai2024aligningtexttoimagediffusionmodels, xu2024imagereward, fan2023dpok,xue2025dancegrpo, lee2023aligning, black2023training}. Reward models, which provide feedback through automated processes or based on human preferences, are increasingly used to evaluate how well the outputs of generative models align with human expectations~\cite{wu2023humanpreferencescorev2, xu2024imagereward}. The feedback scores from these models guide the generative processes toward producing outputs that exhibit higher quality and greater controllability.
Additionally, pre-trained vision models are frequently employed as reward models~\cite{black2024trainingdiffusionmodelsreinforcement, prabhudesai2024aligningtexttoimagediffusionmodels, li2024controlnetimprovingconditionalcontrols}. These models offer more granular and objective control targets, closely approximating human feedback levels. In this paper, AssetDropper utilizes a diffusion-based reward model, which significantly enhances the quality and human-preferred aesthetics of the assets extracted, surpassing the performance of the original model.

\section{Methodology}

\subsection{Preliminaries}
\label{sec:preliminaries}
\paragraph{Latent Diffusion Models (LDMs)} The LDM~\cite{rombach2022high} operates as a two-stage diffusion framework that includes an autoencoder $\mathcal{E}$ and a UNet-based noise predictor. The following formula can represent the optimization process:
\begin{equation}
\mathcal{L}=\mathbb{E}_{\bm z_t,\bm c,t,\bm \epsilon \sim\mathcal{N}(0,I)}(||\bm \epsilon-\bm {\epsilon_\theta}(\bm z_t,t,\bm c)||_2^2),
\label{formular:diffusionloss}
\end{equation}
where $\bm{z}_t$ represents the noisy version of $\bm z_0$, the target data, at timestep $t$. $\bm c$ denotes the condition information, and $\bm {\epsilon_\theta}$ represents the noise prediction with condition $\bm c$.

\paragraph{Virtual Try-On.}
Virtual try-on~\cite{zhu2023tryondiffusion, choi2024improvingdiffusionmodelsauthentic} involves two main components: a garment conditioning module and a person image synthesis module. The garment conditioning module encodes the garment information via simple layers or a parallel UNet branch, gathering the feature for injection. The person image synthesis module generates the final try-on image, with the guidance of human pose and garment feature. Normally the person image synthesis module is built upon UNet architecture and initialized with Stable Diffusion~\cite{rombach2022high, podell2023sdxl} for strong generative priors. The denoising supervision (Formula \ref{formular:diffusionloss}) fine-tune and optimize the main components with the garment image and human information (e.g., human pose, body mask, and masked human image) as condition $\bm c$. The virtual try-on framework, particularly the approach outlined in~\cite{choi2024improvingdiffusionmodelsauthentic}, can be re-targeted for our task and offer a promising foundation. In detail, we can use the garment conditioning module for attending source image information and also optimize SDXL backbone for synthesizing the asset image. We detail our problem formulation and tailored designs in the following sections.

\subsection{Problem Formulation}
\label{sec:problem_formulation}
Given an image with a selected region indicating the asset to extract, we aim to produce a standardized asset with high fidelity and quality. Previous work~\cite{10.1007/978-3-319-46487-9_27} attempted to extract only window assets from urban images using image rectification and segmentation algorithms. However, this method lacks generalizability and generative capability, and thus fails to meet the requirements of our problem setting. A key challenge that remains unaddressed is dealing with noise, primarily from distortion and occlusion, requiring a mapping from noisy images to standardized assets. Learning such a mapping from 2D data alone is highly challenging, and accurate 3D information is often unavailable. To address this, we design a principled solution, aided by data curation and reward-driven optimization to fulfill this task.

\begin{figure}[htbp]
    \centering
    \includegraphics[width=0.45\textwidth]{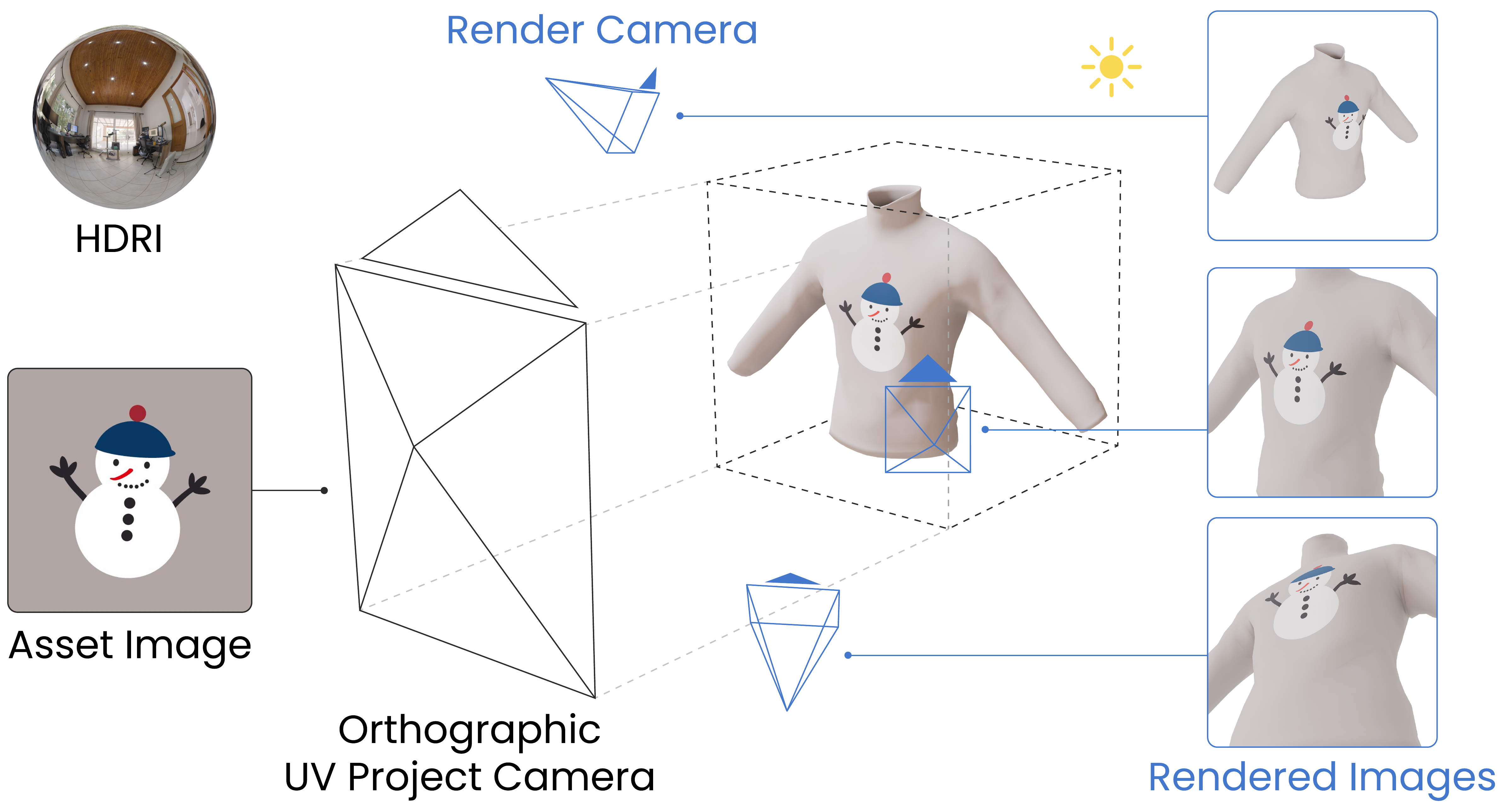}
    \caption{Pipeline of paired training data construction. We use High Dynamic Range Images (HDRI) to simulate the realistic ambient lighting, and then render images from virtual cameras sampled from the front hemisphere.}
    \label{fig:data}
\end{figure}

\subsubsection{Dataset.} Paired data is needed for asset extraction training. Our ultimate goal is to construct paired data consisting of real-world images and their corresponding standardized asset images. To achieve this, we first collected and generated over 10,000 standardized asset images. Then, we created a scene in Blender with realistic environmental lighting through High Dynamic Range Images (HDRI) and imported several commonly seen meshes in daily life. These include meshes with flat surfaces as well as those with curved surfaces of varying degrees. Using UV projection, we mapped the asset images onto these meshes. Finally, we rendered the real-world images from random angles, with the VTON dataset together forming a comprehensive dataset with 200,000 paired samples of real-world images and their corresponding standardized asset images, called the Standardized Asset Palette Dataset (SAP). Fig.~\ref{fig:data} shows this process.

\subsubsection{Benchmark.} Given the lack of quantitative evaluation, it is imperative to establish a benchmark for asset extraction. To ensure reliable assessment, we collected a diverse set of images from open-source platforms and constructed data through synthetic methods, annotated with detailed text instructions and high-quality target masks. To cover different scenarios, our data consists of two types: 1) Real-World VTON dataset (SAP-Real), and 2) Synthetic dataset (SAP-Syn). The resulting SAP benchmark comprises a total of 212,557 image-instruction-mask data samples. The dataset is further divided into two parts: the training set and the test set, which contain 191,301 and 21,256 data samples, respectively. We trained our model on both SAP-Real and SAP-Syn to achieve better robustness. The test set is also split into SAP-Real and SAP-Syn to evaluate the model's performance under different scenarios and conditions.

\subsection{AssetDropperNet}
\label{sec:assetdropper_net}
AssetDropperNet performs asset extraction from a source image with its corresponding mask. Its design is similar to preliminary virtual try-on systems using generative modeling techniques, but they aim at the opposite objectives. We denote $\bm x_\text{r}$ as the reference image of the extracted asset, $\bm m$ as the mask of the asset region which is generated using GroundingDINO~\cite{liu2024groundingdinomarryingdino}, $\bm x_\text{e}$ as the edge map of the extracted asset, and $\bm x_\text{a}$ as the image of the asset. To better leverage the rich prior knowledge of text-to-image diffusion models, we introduce GPT-4o~\cite{gpt4o} to provide captions that describe the details of the assets. Our primary goal is to estimate the asset $\bm x_\text{a}'$ from the reference image $\bm x_\text{r}$. We treat asset extraction as an example-based image inpainting problem~\cite{yang2023paint}. The components include:

\paragraph{FeatureNet.} We use the UNet in Stable Diffusion XL (SDXL)~\cite{podell2023sdxl} as an encoder to extract low-level features of the asset. We concatenate the latent of three inputs, including: 1) $\bm x_\text{r}$, the source image, 2) the resized mask $\bm m$ of the asset region, and 3) the edge map $\bm x_\text{e}$ of the asset. These latents are aligned in spatial dimensions and are concatenated in the channel axis. We inflate the input convolutional layer of the UNet to six channels with a parameter copy. We pass the inputs through the frozen pre-trained UNet encoder to obtain the intermediate representation. 

\paragraph{Image Prompt Adapter.} IP-Adapter~\cite{ye2023ip} is used to capture the high-level semantics of the asset. We extract features using the frozen CLIP image encoder~\cite{ilharco_2021_5143773} and fine-tune feature projection layers and cross-attention layers initialized with a pre-trained IP-Adapter. To accurately extract the features of the asset, we input the masked image $\bm x_\text{m} = \bm m \odot \bm x_\text{r}$ into the IP-Adapter.

\paragraph{ExtractNet.} The SDXL-Inpainting UNet model serves as the main generator, producing the noise prediction of the extracted asset $\bm y$. The features extracted by FeatureNet are integrated with the self-attention layers of ExtractNet, while the cross-attention layers handle features from the IP-Adapter and the text prompt.

In the training stage, the input of the ExtractNet is the latent asset $\mathcal{E}(\bm x_\text{a})$. Therefore, we adjust the convolutional layer of the UNet to four channels with a parameter copy for initialization.

\begin{figure}[htbp]
    \centering
    \includegraphics[width=0.45\textwidth]{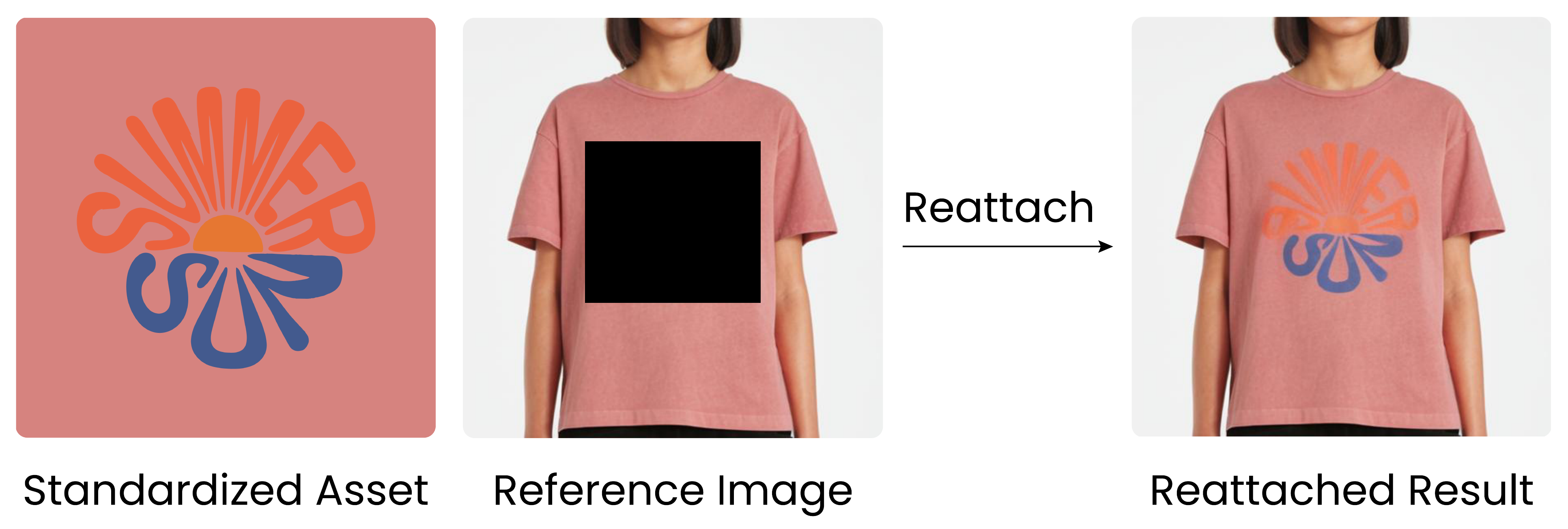}
    \caption{Capability of our reward model. Our reward model reattaches the asset to the reference image within the masked area.}
    \label{fig:inpainting}
\end{figure}

\subsection{Reward-Driven Optimization}
\label{sec:reward_driven_optimization}
\subsubsection{Reward Model Design.} Inspired by ControlNet++~\cite{li2024controlnetimprovingconditionalcontrols}, we aim to design a reward process to better align the generated asset $\bm x_\text{a}$ with the input reference image $\bm x_\text{r}$. Given the absence of a suitable existing model for additional supervision, it is necessary to design and implement a new reward model $\mathcal{R}$. To begin with, an ideal model can be a feed-forward with single-step inference to perform the reverse task of our content extraction. However, both GAN-based and single-step diffusion models fall short in performing the task with robust and high-quality results. Consequently, we train a diffusion-based model tailored to generate the asset in the reference image, akin to virtual try-on.  Although this model achieves satisfactory results, its multi-step inference process renders it impractical for end-to-end training within AssetDropper when used as a reward model. To address this, we employ several estimation techniques to enable feasible training using the inpainting network for reward guidance, which will be discussed in detail later in this section. Fig.~\ref{fig:inpainting} illustrates the effectiveness of our trained reward model in maintaining high fidelity and quality after reattaching the asset.

\begin{figure}[htbp]
    \centering
    \includegraphics[width=0.45\textwidth]{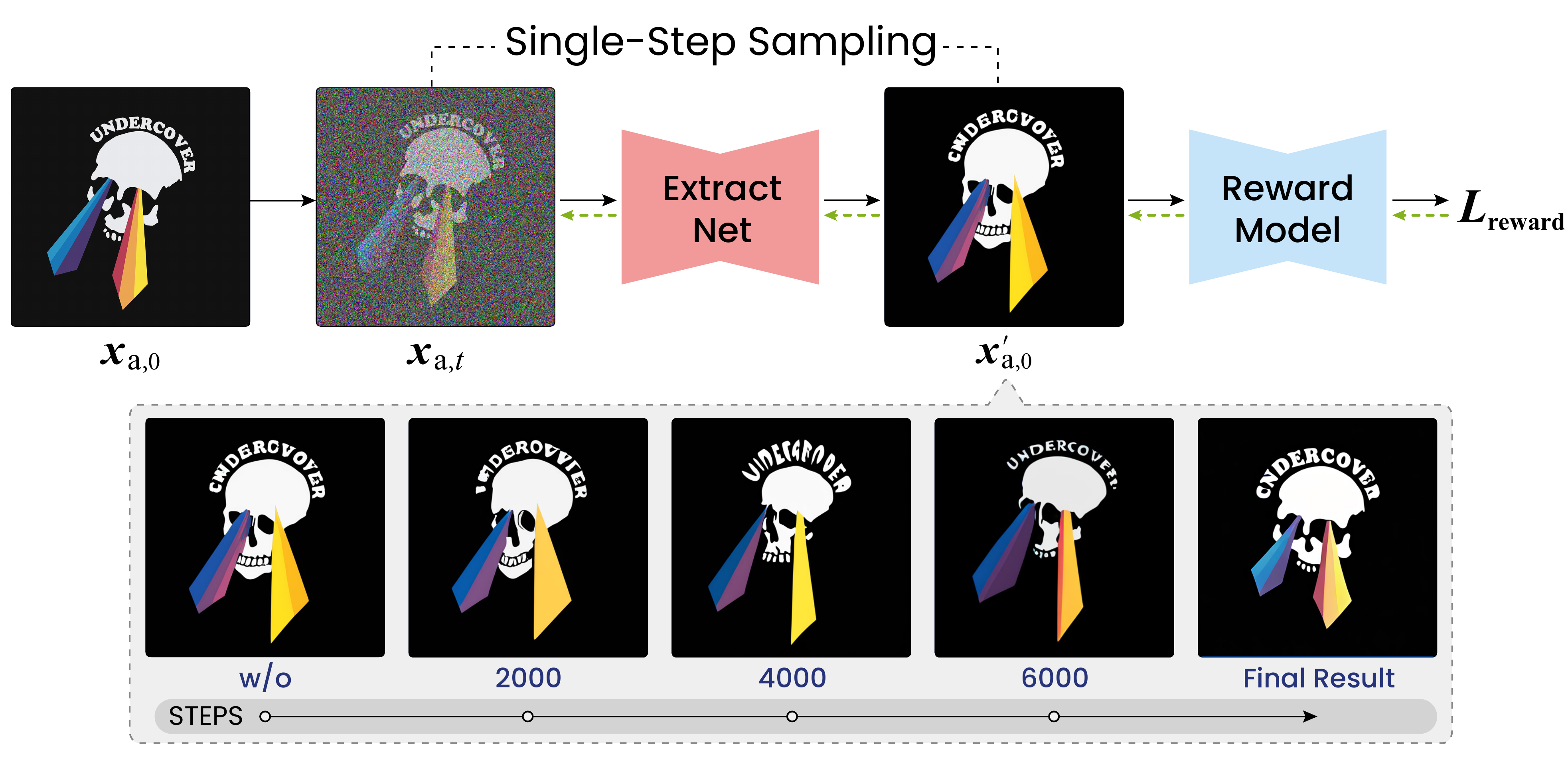}
    \caption{Visualization of the reward fine-tuning effects. The quality of the generated result improves over reward fine-tuning steps.}
    \label{fig:Reward}
\end{figure}

\subsubsection{Efficient Reward Fine-Tuning Strategy.}
To achieve the pixel-space cycle consistency loss, we need to obtain the final output of the diffusion model, which requires multiple denoising steps. This introduces two potential issues: 1) The normal reverse process required in sampling makes training infeasible due to time and GPU memory consumption.
2) The reward model needs to provide effective supervision, indicating that the painting task cannot be overly challenging or simple.
For the first issue, we adopt the same one-step efficient reward strategy as ControlNet++. The prediction of the AssetDropperNet is used to estimate a clean version of the asset image $\bm x'_{\text{a}, 0}$, i.e., the input of the reward model.
Fig.~\ref{fig:Reward} illustrates this process.
We can predict the clean image $\bm x'_{\text{a}, 0}$ by performing single-step sampling given the input, i.e., the ground-truth data with sampled noise $\bm x_{\text{a},t}$:
\begin{equation}
\bm x'_{\text{a}, 0} = \frac{\bm x_{\text{a}, t} - \sqrt{1 - \alpha_t} \, \bm{\epsilon_\theta} (\bm x_{\text{a},t}, \bm x_\text{r}, \bm m, \bm x_\text{e}, \bm c, t)}{\sqrt{\alpha_t}}.
\end{equation}
For the second issue, the key idea is to control the noisy level of the input data, including the asset and the masked reference image. For the estimated clean asset, which is the output of AssetDropperNet, as this estimation is lossy, we only include the reward loss when the timestep $t$ of the AssetDropperNet is small enough, i.e., $t \leq t_{
\text{thres}}$, unless the image is blurry and the task of inpainting is too difficult. For the reference image, as the noise can be controlled to produce the noisy version image of a proper noise level, we select the timestep $t_{\text{c}}$, with trials. As the diffusion-based InpaintNet requires a reverse process for sampling, we also predict the clean reference image $y'_{\text{r},0}$ with estimation by performing single-step sampling:
\begin{equation}
\bm y'_{\text{r},0} = \frac{\bm y_{\text{r},t_\text{c}} - \sqrt{1 - \alpha_{t_{\text{c}}}} \, \bm {\epsilon_{\theta}} (\bm y_{\text{r}, t_\text{c}}, \bm x'_{\text{a},0}, \bm m, \bm c, t_\text{c})}{\sqrt{\alpha_{t_\text{c}}}}.
\end{equation}
We can then calculate the pixel-space cycle consistency loss with the estimated output $\bm y'_{\text{r},0}$ and the original reference image $\bm x_\text{r}$:
\begin{equation}
\mathcal{L}_{\text{reward}}=\mathcal{L}(\bm x_\text{r}, \bm y'_{\text{r},0})=\mathcal{L}(\bm x_\text{r}, \mathcal{R}(\bm x'_{\text{a},0}, \bm y_{\text{r},t_\text{c}})).
\end{equation}
We integrate the reward loss with the original diffusion training loss $\mathcal{L}_{\text{train}}$:
\begin{equation}
    \mathcal{L}_{\text{total}} =
    \begin{cases}
        \mathcal{L}_{\text{train}} + \lambda \cdot \mathcal{L}_{\text{reward}}, & \text{if } t \leq t_{\text{thres}}, \\
        \mathcal{L}_{\text{train}}, & \text{otherwise}.
    \end{cases}
\end{equation}
\section{Experiments}
We evaluate AssetDropper with both synthetic data and real-world images across various scenarios. We introduce the implementation details and experimental setup in Sec.~\ref{Setup} and present the main comparison in Sec.~\ref{sec:Qualitative Comparison} and Sec.~\ref{sec:Quantitative Comparison}. Finally, we show results of ablation studies and user studies in Sec.~\ref{sec:analysis}.

\begin{table*}[t]
\centering
\setlength{\tabcolsep}{20pt}
\caption{Quantitative comparison of different methods evaluated on the SAP-Syn (synthetic data) and SAP-Real (real-world data) datasets. }  
\label{tab:comp}
\begin{tabular}{@{}ccccc@{}}
\toprule
\textbf{Method}            & \textbf{Dataset}    & \textbf{FID}~$\downarrow$    & \textbf{KID}~$\downarrow$   & \textbf{CLIP-I}~$\uparrow$    \\ \midrule
T2I-Adapter              & \multirow{6}{*}{SAP-Syn}       & 106.19   & 0.0274 & 0.9164   \\
InstructPix2Pix         &        & 142.24  & 0.0590  & 0.8966  \\ 
ControlNet        &        & 105.09 & 0.0300 & 0.9141  \\
Ours w/o Reward \& w/o Edge map   &        & 62.41  & 0.0019 & 0.9707  \\
Ours w/o Reward \& w/ Edge map   &        & 60.33  & 0.0017 & 0.9634  \\
\textbf{Ours w/ Reward \& w/ Edge map}   &        & \textbf{50.36}  & \textbf{0.0016}  & \textbf{0.9729}  \\ \midrule
T2I-Adapter            & \multirow{6}{*}{SAP-Real}       & 96.12   & 0.0278 & 0.9164   \\
InstructPix2Pix        &        & 132.75  & 0.0592  & 0.8970  \\ 
ControlNet        &        & 93.68 & 0.0312 & 0.9150  \\
Ours w/o Reward \& w/o Edge map   &        & 49.78  & 0.0015 & 0.9625  \\
Ours w/o Reward \& w/ Edge map   &        & 49.48  & 0.0014 & 0.9639  \\
\textbf{Ours w/ Reward \& w/ Edge map}   &        & \textbf{48.71}  & \textbf{0.0013}  & \textbf{0.9673}  \\ \bottomrule
\end{tabular}
\end{table*}

\subsection{Experimental Setup}
\label{Setup}
\subsubsection{Implementation Details.}
We first train AssetDropperNet and the reward model separately, both using a learning rate of $1\times10^{-6}$ and a batch size of 8 for 120 epochs. It takes ~80 hours to train each model on eight A800 GPUs. When fine-tuning AssetDropperNet with the reward model, we use a learning rate of $1\times10^{-7}$ and a batch size of 1, training for 20,000 steps on eight A800 GPUs in ~2 hours. During reward fine-tuning, we set the $t_{\text{thres}}$ to 300 for AssetDropperNet's input. In addition, we set $\lambda$ to 1.0 and add noise to the reference image input of the reward model with a fixed timestep $t_c = 150$. We achieve the best results by tuning $t_{\text{thres}}$ and $\lambda$. When these values are too small, the reward has no optimization effect; when they are too large, the generated images collapse. We also find that performing some preprocessing on the input images may improves the output quality, such as enlarging the target region or applying super-resolution.

\subsubsection{Baselines.} We compare our method with a wide range of possible solutions for the asset extraction task. An ideal solution would employ universal generative models that follow human instructions, \eg, InstructPix2Pix~\cite{brooks2023instructpix2pix} and OmniGen~\cite{xiao2024omnigen}. We include OmniGen in our comparison to demonstrate the limitations of current omni-generation methods for our task. Another approach, akin to human manipulation processes, involves using novel view generation models such as Zero123++~\cite{shi2023zero123plus} to produce front-view images, coupled with segmentation models like SAM~\cite{kirillov2023segment} to extract the subject. Furthermore, we also compare with other popular Image-to-Image frameworks and fine-tune them for our task. Specifically, we fine-tune InstructPix2Pix~\cite{brooks2023instructpix2pix}, and adapter-based fine-tuned models for controllable generation, ControlNet~\cite{zhang2023adding} and T2I-Adapter~\cite{mou2024t2i}. All three approaches were fine-tuned on the SAP training dataset using the pre-trained weights of SDXL for a fair comparison, with a consistent image resolution of $512\times512$ for all output.

\begin{figure}
    \centering
    \includegraphics[width=0.45\textwidth]{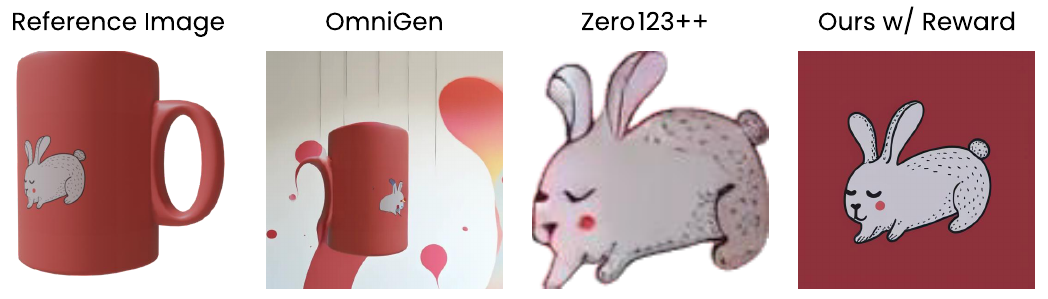}
    \caption{Qualitative comparison with OmniGen and Zero123++. The prompt used in OmniGen is ``The pattern in <img><|Reference Image|></img> is painted on a whiteboard.'' We use Zero123++ to generate multiple additional perspectives of the reference image and select the frontal view. Then, we utilize SAM to extract the asset portions from the results of Zero123++ to obtain the visualization results.}\label{fig:quali_additional}
\end{figure}

\subsubsection{Evaluation Metrics.} To perform a comprehensive quantitative evaluation, we assessed high-level semantic similarity between generated images and reference images by utilizing the CLIP image similarity score~\cite{radford2021learning}. Additionally, to rigorously evaluate the photorealism of the generated images, we computed the Fréchet Inception Distance (FID)~\cite{heusel2017gans} and Kernel Inception Distance (KID)~\cite{binkowski2018demystifying} scores. The metric measures FID on the $299\times299$ center-cropped patches of each image to evaluate high-resolution details. We calculate these metrics on 1,000 samples from both the SAP-Syn and SAP-Real test datasets. These scores indicate the visual quality and diversity of generated images compared to real-world distributions.

\subsection{Qualitative Comparison}
\label{sec:Qualitative Comparison}
We compare our method with the general visual generation model OmniGen~\cite{xiao2024omnigen} and the novel view synthesis model Zero123++~\cite{shi2023zero123plus}. As shown in Fig.~\ref{fig:quali_additional}, OmniGen can follow editing instructions but struggles with fine-grained comprehension, resulting in inaccuracies in asset extraction. Zero123++, on the other hand, suffers from low resolution in novel view generation, leading to blurrier outputs when SAM is used for asset extraction.

As shown in Fig.~\ref{fig:quali_real} and Fig.~\ref{fig:quali_syn}, we compare the visualization results of our method with other approaches to the SAP-Syn and SAP-Real datasets. T2I-Adapter captures asset outlines but struggles with color accuracy and structural consistency, leading to distortions. ControlNet achieves better edge precision but fails to reproduce vibrant colors and fine details. InstructPix2Pix performs poorly in this task, producing blurry, distorted results and unwanted textures outside the asset. In contrast, our method without the reward mechanism better preserves overall shape and structure but lacks color fidelity and detail. With the reward mechanism, our method significantly improves in both color and detail restoration, outperforming all baselines and demonstrating the effectiveness of our reward design. Furthermore, our approach generalizes well to generated images, as shown on 2D and multiview 3D data from FLUX~\cite{flux2023} and MV-Adapter~\cite{huang2024mvadaptermultiviewconsistentimage}.

\subsection{Quantitative Comparison}
\label{sec:Quantitative Comparison}
As shown in Tab.~\ref{tab:comp}, our method outperforms other image-to-image methods across all evaluation metrics. For the SAP-Syn dataset, our method achieves an FID score of 60.33 without reward and 50.36 with reward, significantly lower than T2I-Adapter, InstructPix2Pix, and ControlNet. Similarly, for the SAP-Real dataset, our method achieves an FID score of 49.48 without reward and 48.71 with reward, again outperforming the other methods. These results demonstrate the superiority of our method in generating high-quality images with better alignment to the target attributes.

\begin{figure}[htbp]
    \centering
    \includegraphics[width=0.45\textwidth]{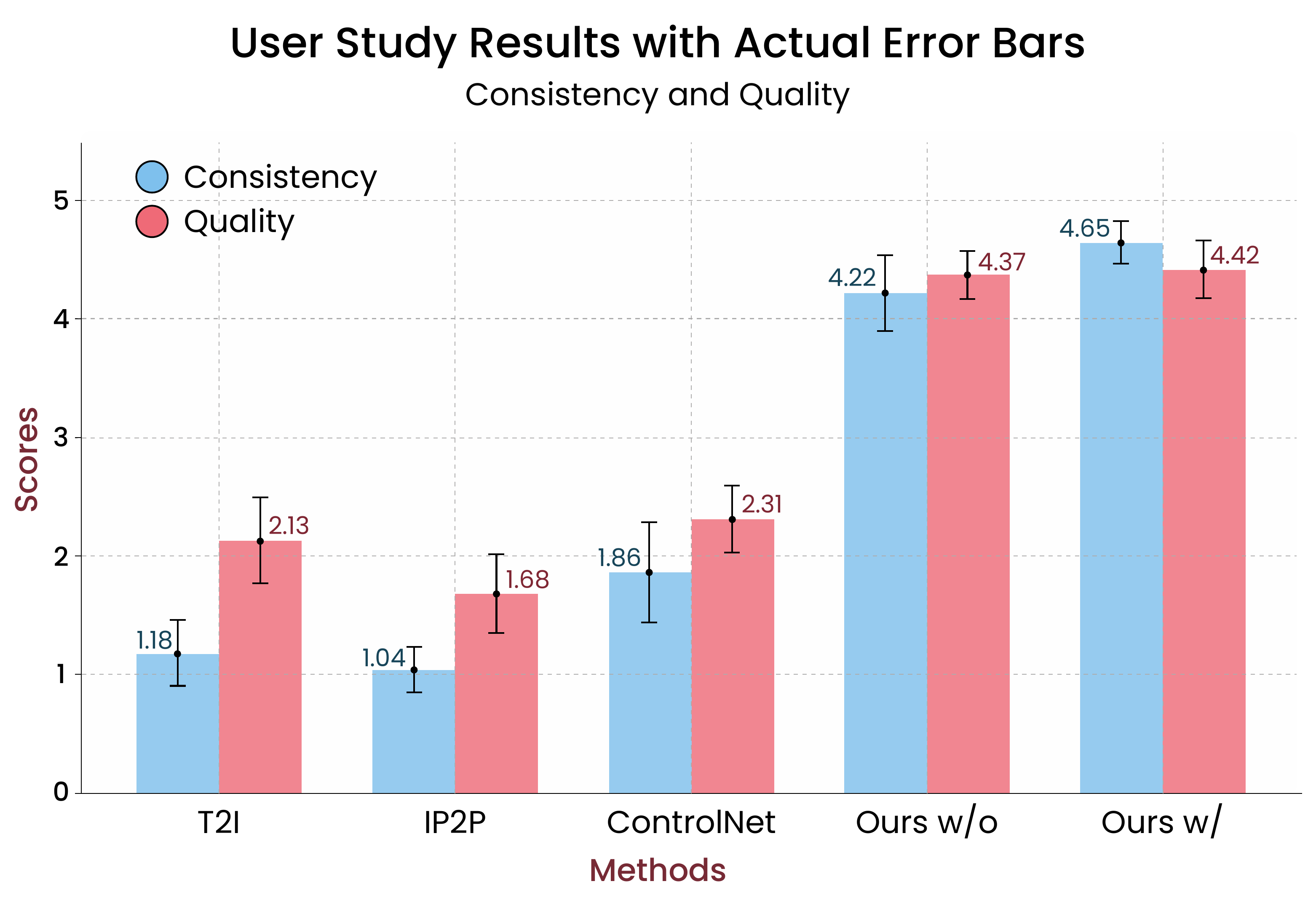}
    \caption{Result of our user study. Among all methods, ours w/ achieves the highest scores in both consistency and quality, demonstrating its superior overall performance.}
    \label{fig:userstudies}
\end{figure}

\begin{figure}
    \centering
    \includegraphics[width=0.45\textwidth]{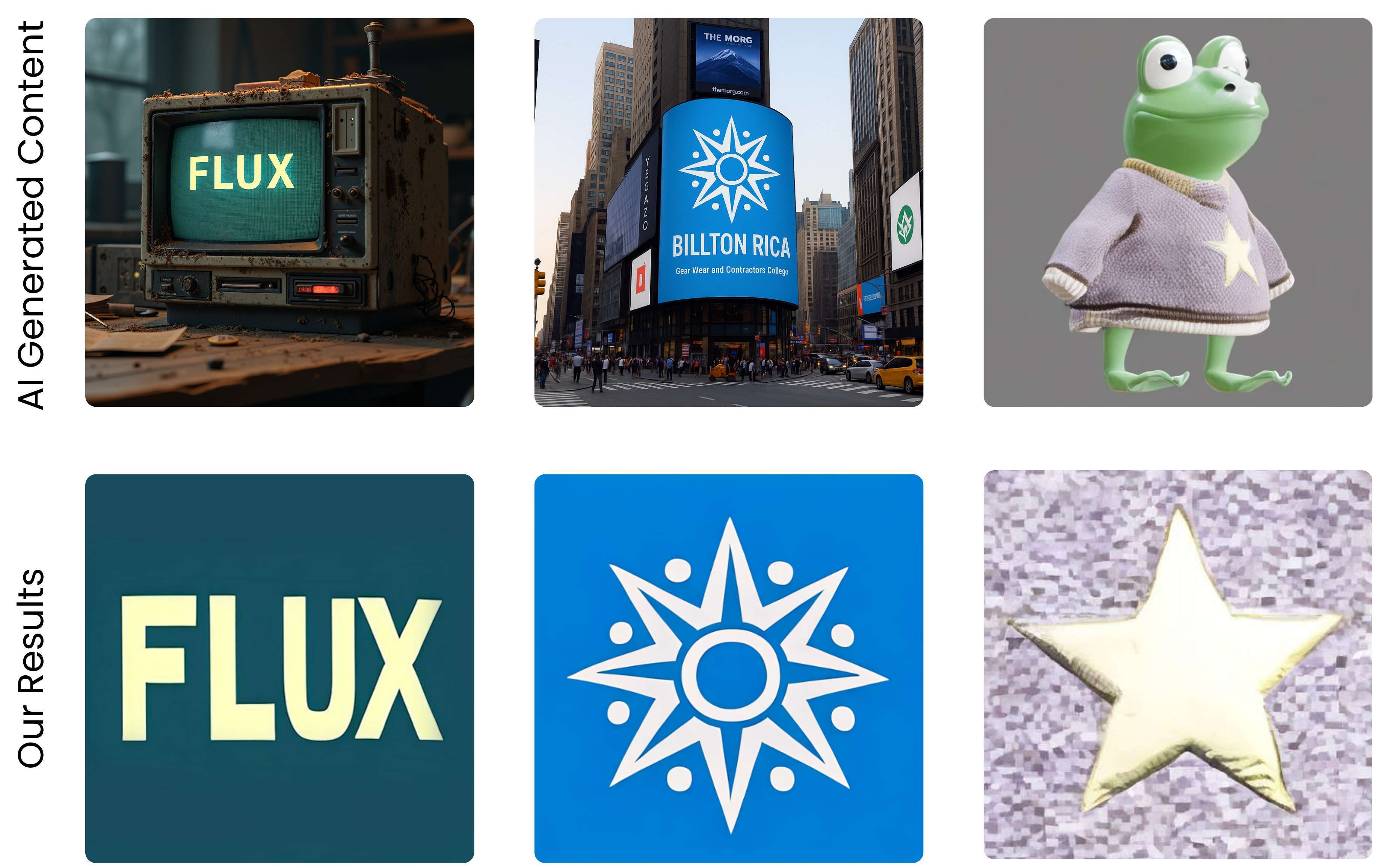}
    \caption{Applications of AssetDropper across different scenarios. The first two images are 2D images generated by FLUX, while the third image is one view selected from 3D multi-view images generated by MV-Adapter. In these scenarios, our model can still complete the task with high quality.}
    \label{fig:quali_aigc}
\end{figure}

\subsection{Analysis}
\label{sec:analysis}

\subsubsection{Ablation Study.}
\begin{figure*}[t]
    \centering
    \includegraphics[width=\textwidth]{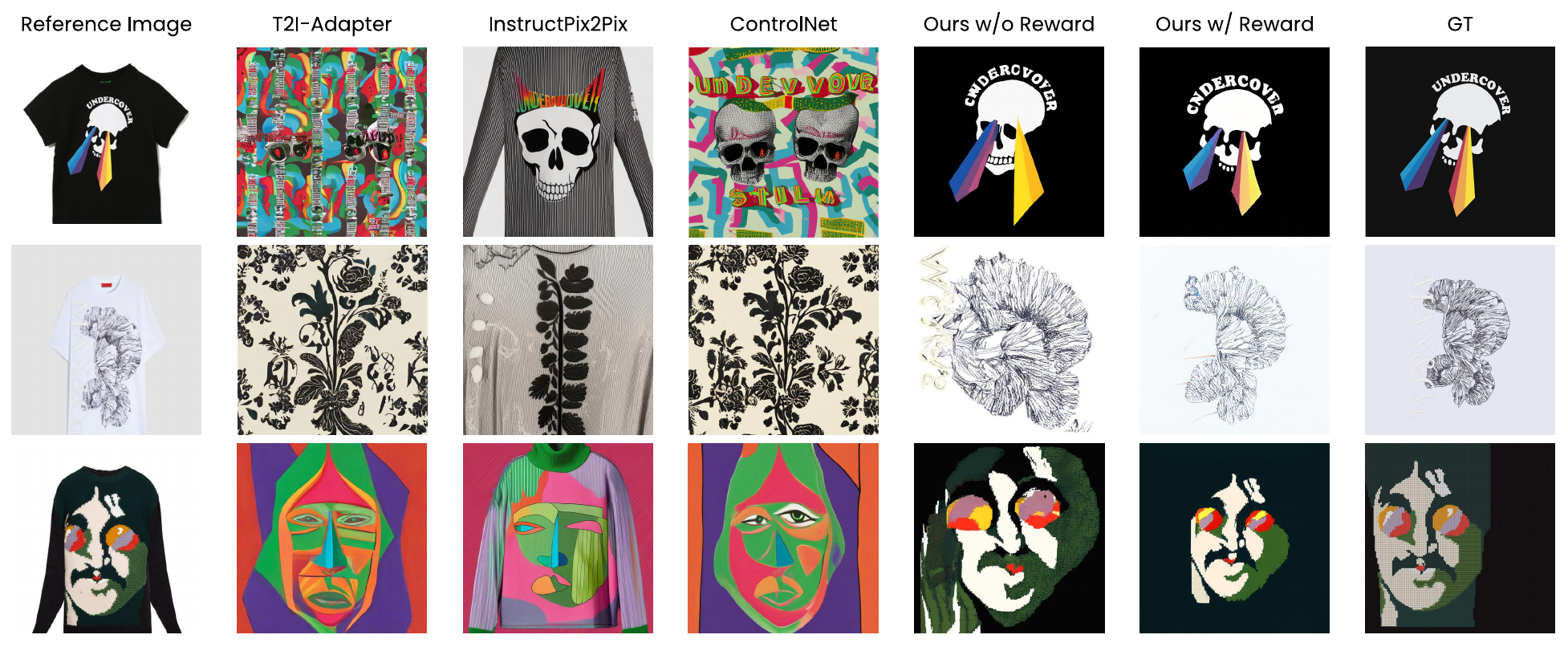}
    \caption{Qualitative results in SAP real-world test dataset. Other methods can only roughly preserve the semantics of the reference image but fail to maintain consistency. In contrast, our method maintains a high level of consistency, achieving the goal of asset extraction.}
    \label{fig:quali_real}
\end{figure*}

\begin{figure*}[t]
    \centering
    \includegraphics[width=\textwidth]{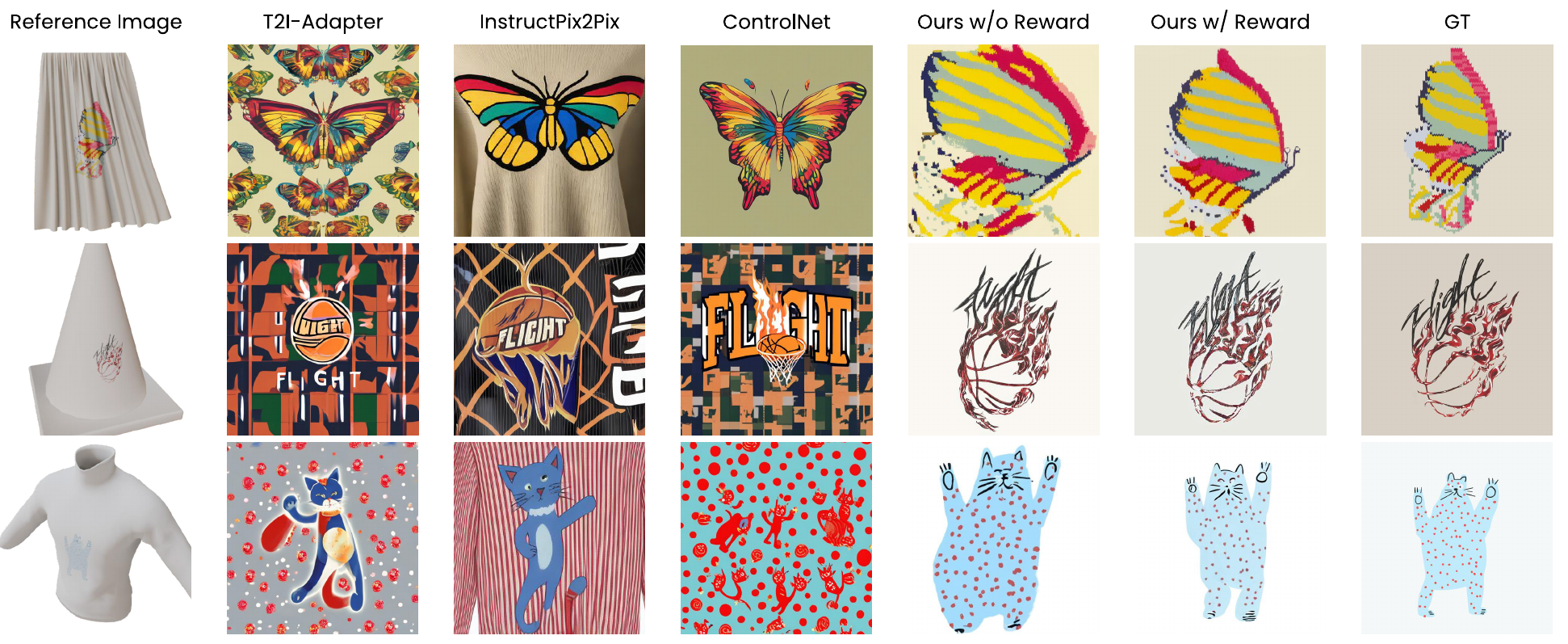}
    \caption{Qualitative comparison in SAP synthetic test dataset. For the synthetic test dataset, we selected three different meshes corresponding to three varying levels of surface curvature. Similar as the results on SAP-Real, our method achieves asset extraction results that are close to the ground truth.}
    \label{fig:quali_syn}
\end{figure*}

We conduct an ablation study to evaluate the impact of the reward model and the use of edge maps in AssetDropperNet. We adopt FID scores, KID scores, and image similarity scores as evaluation metrics. As shown in Tab.~\ref{tab:comp}, incorporating edge maps during training and applying the reward strategy in post-training both lead to noticeable improvements in performance. Fig.~\ref{fig:quali_real} and Fig.~\ref{fig:quali_syn} present the qualitative results with and without the reward model, further demonstrating that the reward model plays a critical role in enhancing the performance of AssetDropperNet, particularly in preserving the fidelity of complex assets.

\subsubsection{User Study.}
To evaluate the performance of our proposed method, we conduct a comprehensive user study focusing on two key aspects: 1) the consistency between the generated images and the reference images, and 2) the perceived quality of the generated images.
We recruited 25 participants for the user study. Each participant evaluated the outputs of different methods on a shared set of 10 image groups, where higher scores indicate better performance.
As shown in Fig.~\ref{fig:userstudies}, our model achieves much higher user preferences in maintaining the consistency and quality of generated assets compared to baseline approaches.

\section{CONCLUSION}
This paper has presented AssetDropper, a novel diffusion model framework for asset extraction, particularly in in-the-wild scenarios. To tackle challenges like generalization, distortions, and occlusions, we incorporate two key components: dataset curation and reward-driven optimization, both tailored for asset extraction. Extensive experiments across diverse datasets demonstrate that our approach preserves asset details and generates high-fidelity images. We further showcase its potential in real-world settings, highlighting its practical applicability and effectiveness. However, our model may struggle with severe occlusions or views where excessive information is lost.
Future work includes leveraging temporal priors from video models to further enhance the quality and investigating broader applications.

\begin{acks}
We thank Qianxi Liu for assisting with figure creation and for the valuable feedback. We thank Yifu Technology (Guangzhou) Co., Ltd for providing part of the dataset.
This work was partially supported by Guangzhou Basic Research Scheme \#2024A04J4229, Guangzhou Industrial Information and Intelligent Key Laboratory Project \#2024A03J0628, the Research Grants Council of Hong Kong (C5055-24G and T45-401/22-N), and the National Natural Science Foundation of China (No. 62201483).
\end{acks}

\bibliographystyle{ACM-Reference-Format}
\bibliography{bibliography}

\clearpage

\begin{figure*}[t]
   \centering
   \includegraphics[width=0.8\textwidth]{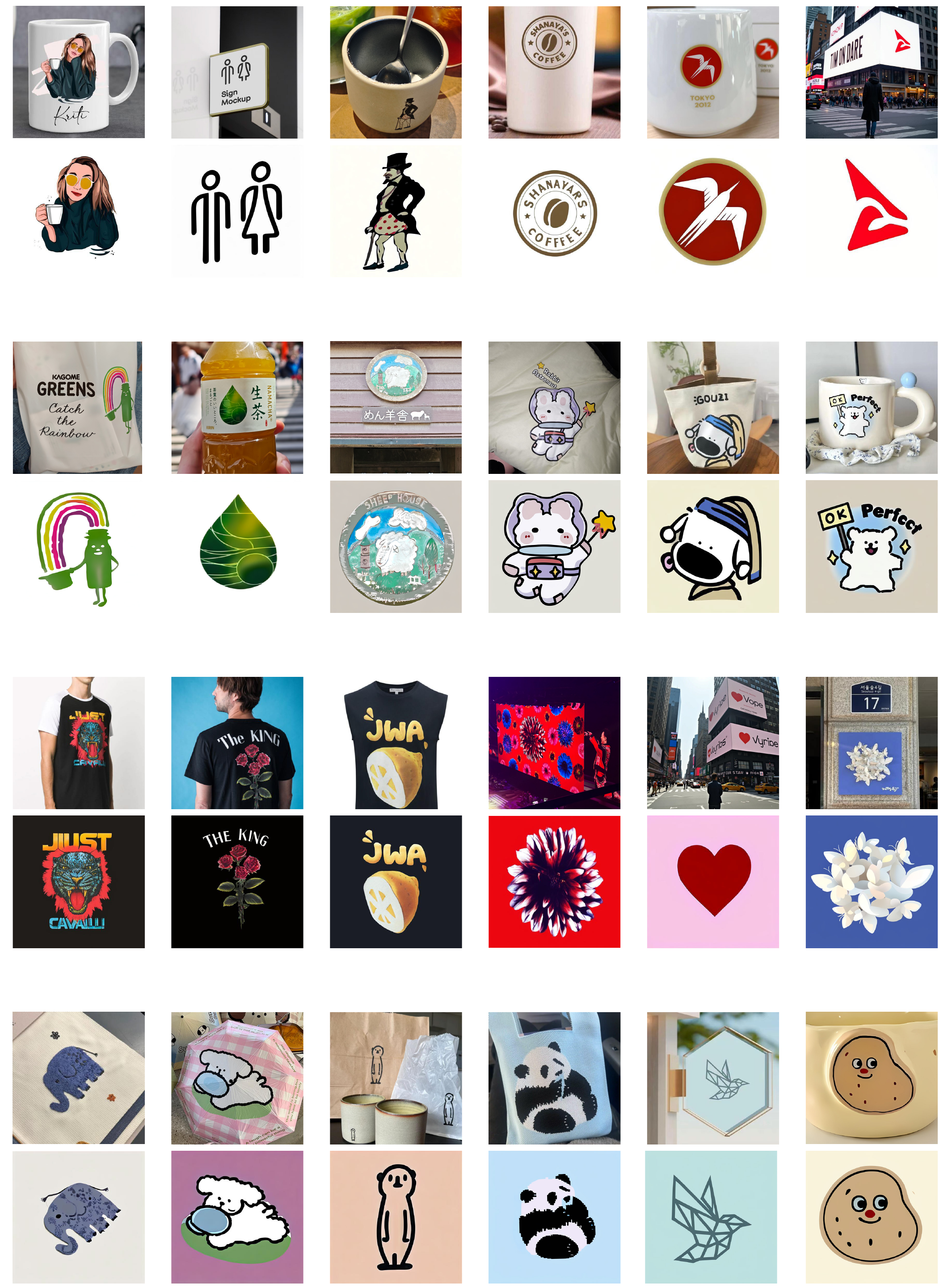}
   \caption{Qualitative results on in-the-wild images. For each image block, the first row is the input reference image, and the second row is the output of AssetDropper.}\label{fig:real_test1}
\end{figure*}

\end{document}